# Learning From Positive and Unlabeled Data Using Observer-GAN

**Omar Zamzam, Haleh Akrami, Richard M. Leahy**

Signal and Image Processing Institute, University of Southern California, Los Angeles, USA

## Abstract

The problem of learning from positive and unlabeled data (A.K.A. PU learning) has been studied in a binary (i.e., positive versus negative) classification setting, where the input data consist of (1) observations from the positive class and their corresponding labels, (2) unlabeled observations from both positive and negative classes. Generative Adversarial Networks (GANs) have been used to reduce the problem to the supervised setting with the advantage that supervised learning has state-of-the-art accuracy in classification tasks. In order to generate *pseudo*-negative observations, GANs are trained on positive and unlabeled observations with a modified loss. Using both positive and *pseudo*-negative observations leads to a supervised learning setting. The generation of pseudo-negative observations that are realistic enough to replace missing negative class samples is a bottleneck for current GAN-based algorithms. By including an additional classifier into the GAN architecture, we provide a novel GAN-based approach. In our suggested method, the GAN discriminator instructs the generator only to produce samples that fall into the unlabeled data distribution, while a second classifier (observer) network monitors the GAN training to: (i) prevent the generated samples from falling into the positive distribution; and (ii) learn the features that are the key distinction between the positive and negative observations. Experiments on four image datasets demonstrate that our trained observer network performs better than existing techniques in discriminating between real unseen positive and negative samples.

## Introduction

In real-world binary classification problems, it is not unusual to find the cost of labeling one of the classes considerably higher than the cost of labeling the other class, it is sometimes even impossible to label a subset of the data that can belong to either of the classes. This results in an abundance of data for which we have no knowledge of the associated class and a subset of data with labels corresponding to only one class (here we will assume such labels are 'positive'). This is refered to as the problem of PU Learning, or learning from positive and unlabeled input (Bekker and Davis 2020). A common example of the PU data setting is in recommendation systems, where previous purchases or user clicks are direct indicators of user interest (positive label), while labels for all other instances remain unknown (unlabeled)(Zhou et al. 2021). Another important example of PU datasets is in automatic diagnostic systems, where a specific symptom of a disease appears in only a subgroup of patients; the presence of this symptom can be used to label positive label cases, but the absence of this symptom does not necessarily indicate absence of the disease (Claesen et al. 2015). PU learning also has applications in the field of gene identification (Mordelet and Vert 2011), matrix completion (Hsieh, Natarajan, and Dhillon 2015), clustering (Zhou et al. 2009), and spam detection (Wu et al. 2018).

In the current study, we provide a technique that uses a Generative Adversarial Network (GAN) (Goodfellow et al. 2014) to detect features from the unlabeled data for distinguishing between positive and negative classes. The main idea of a GAN is to make the generator network learn the distribution of a given training set so that it can generate new samples from the same distribution. This is achieved by training a generator and a discriminator network adversarially, i.e. the discriminator tries to identify a data sample as real (coming from the training set) or fake (generated by the generator network). The objective of the generator network is to generate data samples that the discriminator fails to classify as fake. We use the unlabeled dataset in this context as the input training set to the discriminator. In addition, we use an additional classifier, here called an *observer network*, to determine whether a data sample comes from the positive set or is fake (generated by the generator network). The objective of our generator network is modified to generate data samples that: (i) the discriminator network fails to classify as fake, and (ii) the observer network successfully identifies as being "negative". These samples should then be generated exclusively from the negative distribution in the unlabeled data. The discriminator network cannot determine that these generated samples are fake, and the observer network can separate them from the positive samples. After training these three networks iteratively, the observer network learns features that are sufficient to classify between unseen positive and negative samples.

Nguyen et al. (2017) have previously suggested training a GAN with two discriminators in order to enhance its performance and prevent mode collapse. Their work is based on one discriminator ($D_1$) that favors real data samples, and a second discriminator ($D_2$) that favors generated samples. The network $D_2$ ensures that the generated samples are sufficiently diverse and different from the samples in the training set and helps to address the mode collapse problem. We adopt a similar approach, with the modification that the training set feeding $D_1$ (named $D$ in this paper), is different from the training set feeding $D_2$ (named $Observer$ in this paper). The unlabeled set is used for training $D_1$ and the positive set is used for training $D_2$. This novel setting allows the generator network to generate samples that are similar to the unlabeled set yet are sufficiently different from the positive set that they are comparable to the negative subset of the unlabeled data. While we have found that a GAN learns more slowly to generate realistic-looking images when trained with two discriminators, we demonstrate that the $Observer$ network can learn to distinguish between the positive and negative distributions without having to generate realistic-looking images.

We evaluate the proposed method, *Observer-GAN*, on four image datasets and show that it outperforms state-of-the-art PU learning methods on binary classification tasks.

## Related Work

The PU learning problem has been investigated since at least 1998 (Comité et al. 1999). Recently there has been a growing interest in developing PU learning methods as a result of the increased use of machine learning/deep learning in a variety of applications where the cost of labeling is high, such as in medicine, marketing, and advertising ((Liu et al. 2003), (Yu, Han, and Chang 2004), (Zhang and Lee 2005), (Elkan and Noto 2008), (Zhou et al. 2009) (Hsieh, Natarajan, and Dhillon 2015), (Chiaroni et al. 2020),(Garg et al. 2021), (Zhao et al. 2022)).

One of the common approaches in PU learning leverages biased Learning, in which unlabeled examples are considered negative examples with label noise, and a binary classifier is trained using a biased cost function that assigns a higher penalty for misclassification of positive examples (clean labels) (Liu et al. 2003), (Hsieh, Natarajan, and Dhillon 2015), (Mordelet and Vert 2014). Another class of methods for PU learning assumes a positive class prior ($P[y = 1]$) is known, which facilitates the training and tuning of a binary classifier. Using this information, training can be stopped when the proportion of identified positive examples in an unlabeled validation set is equal to the positive class prior (Kiryo et al. 2017), (Zhao et al. 2022), (Plessis, Niu, and Sugiyama 2015). However, the true class prior is usually difficult to obtain and so an alternative is to estimate the positive class prior as a first step and then train a binary classifier using this estimated prior (Ivanov 2020). Convergence can be achieved by iterating between these two steps ($TED^n$) (Garg et al. 2021) .

Another common approach for PU learning uses a distance metric to find reliable negative examples which simplifies the problem to the supervised setting. Reliable negative examples are the unlabeled examples that are most different from the positive samples (Yu, Han, and Chang 2004) (Grinenko et al. 2018). To generate negative samples a generative model can also be used (Chiaroni et al. 2020). In the first step, the so-called *D-GAN* is trained with a generator network together with a discriminator network that takes both unlabeled and positive data as training set. The function of this discriminator is to force the generator network to learn to generate pseudo-negative samples. This discriminator network is trained to classify between the unlabeled data as one class, and both the positive data and generated data as another class. Using this setup, the generator learns to fool the discriminator by generating only negative samples. In the second step, a binary classifier is trained given the positive samples and the generated "negative" samples. This setup can work if two requirements are satisfied: (1) The discriminator network has to learn the difference between the unlabeled and the positive sets. This is usually hard to accomplish if the labeled positive samples are Selected completely at random (SCAR (Elkan and Noto 2008)) from the whole positive class distribution. Most real-word PU datasets typically reflect this pattern. Also, Elkan and Noto (2008) show that learning a classifier that can differentiate between positive and unlabeled data is enough to learn the sought-after positive versus negative classifier, knowing the positive class prior, which can sometimes be available or estimated from the data; (2) very realistic generated negative samples are needed to train the second-step classifier. We have observed that this is a bottleneck when the target dataset is complicated and difficult for a GAN to generate. In these cases, the second-step classifier only learns the difference between real and fake images, as this task is easier to learn than the rea-positive versus real-negative sample classification.The learned features that relate to the generated negative samples are therefore probably not generalizable to real negative samples. This difference cannot be ignored by the classifier, and as a result, when used on unseen real positive and negative samples, the network tends to decide that all samples are real.

Our suggested framework can address this problem by easing the above constraints. Although we are generating negative samples, neither of the mentioned conditions are required to achieve the positive-versus-negative classification. The objective of our discriminator is to classify between generated samples and unlabeled samples, and the objective of our $Observer$ network is to classify between generated samples and negative samples. These objective can are achievable even when the SCAR assumption holds. The objective of the generator network is to generate samples that can fool the discriminator, while being identifiable by the observer network as (non-positive) samples. Training the three networks iteratively, drives the $Observer$ network to only learn the features that are representative of the positive-versus-negative difference as described in more detail below.

## Problem Setup

We denote an unlabeled observation as $x_U \in X_U \sim P_U$, such that $X_U$ is the unlabeled random variable and $P_U$ is the unlabeled data distribution. We defined the positive data sample $x_P \in X_P \sim P_P$, such that $X_P$ is the positive random variable and $P_P$ is the positive data distribution.

$z \sim P_Z$ is a random noise vector where $P_Z = N(0,1)$ is a normal distribution. $x_Z$ is a generated sample obtained from the generator. We denote the label for each observation as $y \in \{0, 1\}$. $D(\cdot)$ is the output probability of the discriminator, $G(z)$ is the output sample of the generator network, and $Ob(\cdot)$ is the output probability of the observer network. $\alpha$ is the proportion of the positive samples in the unlabeled dataset, i.e., $P_U = \alpha P_P + (1 - \alpha)P_N$, where $P_N$ is the negative data distribution.

We aim to learn $f(x) = p(y = 1 \mid x)$, a classifier that estimate the true label of an observation $x$. We claim that at the end of the training of our *Observer-GAN*, we have $Ob(x) \approx f(x)$.

## Proposed Method

The standard GAN (Goodfellow et al. 2014) discriminator network ($L_D$) and the generator network ($L_G$) loss functions are as follows:

$$L_D = E_{x_U \sim P_U}[H(D(x_U), 1)] + E_{z \sim P_Z}[H(D(G(z)), 0)] \quad (1)$$

$$L_G = E_{z \sim P_Z}[H(D(G(z)), 1)] \quad (2)$$

where $H(\hat{y}, y) = -y\log(\hat{y}) - (1-y)\log(1-\hat{y})$ is the binary cross entropy loss between $\hat{y}$ and $y$. Minimizing the loss in (1) makes the discriminator able to learn the difference between the data in $P_U$ and the output of the generator $G(z)$. While minimizing the loss in (2) attempts to fool the classifier into believing its samples are real. Minimizing the two loss functions iteratively, results in a generator that is able to generate new samples that are indistinguishable from the real data samples. Our *Observer-GAN* has the same objective for the discriminator, where the input data samples to the discriminator come from the unlabeled dataset $X_U$, so that the generator learns to generate samples that are indistinguishable from the unlabeled data. In addition to this setup, as illustrated in Figure 1, we added a second classifier network (the Observer). The loss function of the Observer network $L_{Ob}$ is defined as follows:

$$L_{Ob} = E_{x_P \sim P_P}[H(Ob(x_P), 0)] + E_{z \sim P_Z}[H(Ob(G(z)), 1)] \quad (3)$$

Minimizing this loss forces the observer network to learn features that separate the positive dataset and the generated samples. We refer to these features as "positive features". We also replace the loss function of the generator network defined in (2) to:

$$L_G = E_{z \sim P_Z}[H(D(G(z)), 1)] + E_{z \sim P_Z}[H(Ob(G(z)), 1)] \quad (4)$$

The first term in (4) is the same as the first term in (2) while the second term in (4) ensures that the samples that the generator produce are different from positive samples and this difference can be learned by the observer. The generator in this setup attempts to fool the discriminator network, while keeping the loss of the observer network minimal. In this way, the generator network captures the distribution of the samples that are presented in the unlabeled dataset (to fool the discriminator), and absent in the positive dataset (to keep $L_{Ob}$ minimal). Minimizing (1), (3), and (4) iteratively, ensures that the observer network learns appropriate features from these generated samples to distinguish between positive and generated samples. Algorithm 1 shows the steps for training the three networks.

---

**Algorithm 1: Training Observed-GAN**

---

1: Initialize network weights $\theta_D$, $\theta_{Ob}$, and $\theta_G$ randomly.
2: **for** number of epochs **do**
3:     **if** epoch number = 100*k for $k \in \{1, 2, 3, ...\}$ **then**
       Reinitialize network weights $\theta_{Ob}$.
4:     **end if**
    Sample $x_U = [x_U^1, x_U^2, x_U^3, ..., x_U^k]$ from $X_U$.
    Sample $x_P = [x_P^1, x_P^2, x_P^3, ..., x_P^k]$ from $X_P$.
    Sample $z = [z^1, z^2, z^3, ..., z^k]$ from $P_Z$.
    Pass $z$ through G to get $x_z = [x_z^1, x_z^2, x_z^3, ..., x_z^k]$
    Calculate the gradient of $L_D$:
    $\nabla_{\theta_D} \frac{1}{k} \sum_{i=1}^{k}[H(D(x_U^i), 1) + H(D(x_z^i), 0)]$
    Use the stochastic gradient of $L_D$ to update D.
    Calculate the gradient of $L_{Ob}$:
    $\nabla_{\theta_{Ob}} \frac{1}{k} \sum_{i=1}^{k}[H(Ob(x_P^i), 0) + H(Ob(x_z^i), 1)]$
    Use the stochastic gradient of $L_{Ob}$ to update Ob.
    Calculate the gradient of $L_G$:
    $\nabla_{\theta_G} \frac{1}{k} \sum_{i=1}^{k}[H(D(x_z^i), 1)] + [H(Ob(x_z^i), 1)]$
    Use the stochastic gradient of $L_G$ to update G.
5: **end for**

---

This training strategy forces the generator network to generate new samples that are closer in distribution to the unlabeled samples, while keeping the features that the observer network did not previously identify as "positive features". The persistence of these features in the generated samples also forces the observer network to only pay attention to the persistently generated negative features, as the rest of the features in the generated samples change more drastically as training progresses.

In the first training steps of *Observer-GAN*, the generated samples are very different from the positive data samples, and since the generator and observer are not training adversarially, the value of $L_{Ob}$ is always small, hence, the second term in (4) is almost disabled. As the training proceeds, when the generator starts learning to generate samples that are closer to samples of the unlabeled dataset (by minimizing the first term in (4)), the value of $L_{Ob}$ increases for the samples $x_Z$ that are closer to the positive dataset, penalizing the generator for generating these samples, We chose the "Observer" because of this behaviour of the network.

This training procedure can result in an undesirable overfitting issue, in which the observer network starts to memorize positive samples. This is due to the fact that the positive training set typically has a small size compared to the unlabeled dataset. In that case, the value of $L_{Ob}$ never increases, and never affects the learning of the generator. To address this issue and avoid this undesirable overfitting, we randomly re-initialize the observer network weights every 100 training epoch. This step has the specific purpose of removing any memory the observer network has about positive samples, so that it always learns meaningful features from the input samples, instead of simply memorizing positive samples. This reinitialization trick has previously been used to enhance GAN performance (Wang et al. 2021). Since the idea of using a validation set to monitor the training and avoid overfitting cannot be used in a PU data setting, this reinitialization step represents a solution to ensure the validity of the final classifier. The observed continuous improvement of the quality of the generated samples, justifies use of reinitialization.

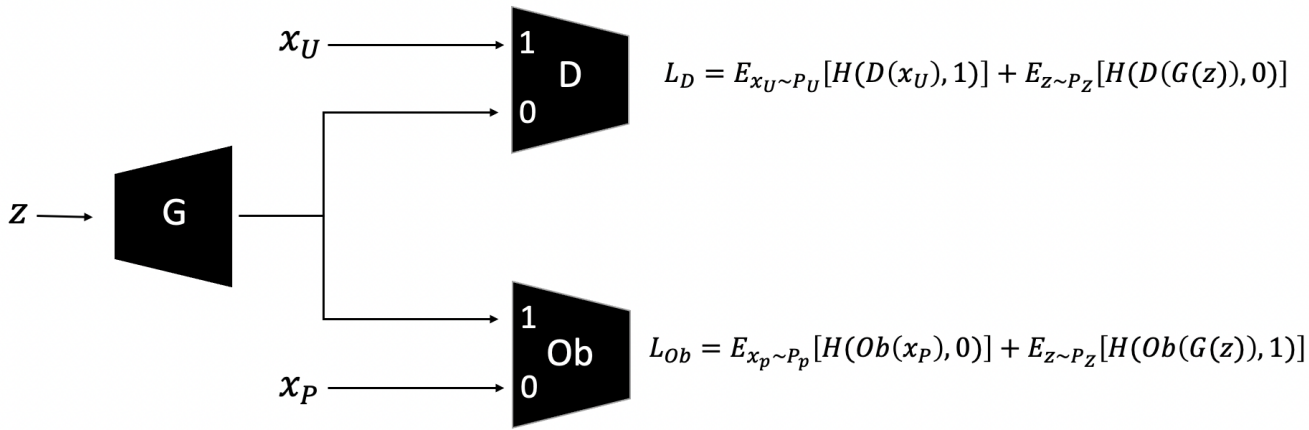

Figure 1: Illustrative figure of *Observer-GAN*

## Results

### Data Preparation

We evaluate the performance of the *Observer* network on four different datasets: MNIST (Deng 2012), Fashion-MNIST (Xiao, Rasul, and Vollgraf 2017), CIFAR-10 (Krizhevsky, Hinton et al. 2009), and animal faces (AFHQ) (Choi et al. 2020). The positive and negative classes are defined respectively as: even versus odd digits on MNIST dataset, last five classes vs first five classes on Fashion-MNIST (classes: T-shirt, Trouser, Pullover, Dress, Coat, Sandal, Shirt, Sneaker, Bag, and Ankle boot), animal versus not animal images on CIFAR-10, and cat versus dog images on AFHQ.

We define the whole training dataset to be $X = \{x_1, x_2, ..., x_p, x_{p+1}, x_{p+2}, ..., x_{p+n}\}$, where $p$ is the number of positive samples, and $n$ is the number of negative samples. We randomly select $\alpha|X_U|$ positive samples to be in the unlabeled set, where $\alpha$ is the proportion of positive samples in $X_U$, and $|X_U|$ is the size of $X_U$. The remainder $(1-\alpha)|X_U|$ samples in $X_U$ are negative. We use $\alpha = 0.5$ in the following experiments. Table 1 shows the number of samples used for each dataset. In all test sets, the number of samples in the positive class is equal to the number of samples in the negative class.

Table 1: Number of samples in each dataset

| Dataset | $X_U$ size | $X_P$ size | Test set size |
|---|---|---|---|
| Fashion-MNIST | 19000 | 19000 | 10000 |
| MNIST | 19000 | 19000 | 10000 |
| AFHQ | 3300 | 3300 | 1000 |
| CIFAR10 | 16000 | 16000 | 10000 |

### Model Architectures

We use convolutional neural networks for all of the datasets. Network architectures are inspired by (Chiaroni et al. 2020). The size of the images in MNIST and CIFAR-10 datasets are $28 \times 28$ and $32 \times 32$ respectively. Figure 2 shows the architectures used for both datasets. We use a Minibatch size of 64 for both datasets.

For the AFHQ dataset, we use input images of size $64 \times 64$, and a minibatch size of 16. Figure 3 shows the network architecture used for training on AFHQ. We use ADAM with a learning rate of 0.0002 for all of the experiments.

We compared our method with two state of the art approaches, $TED^n$ (Garg et al. 2021) and *D-GAN* (Chiaroni et al. 2020). We ran $TED^n$ based on their released implementation[1], which contains three different architectures, we chose the best performing one for each of the datasets. To implement *D-GAN*, we follow the model architectures the authors propose (which are also the base architectures for our method). For the second stage classifier we use the

---
[1]$https://github.com/acmi-lab/PU_learning$

|  |
|---|
| **Input:** $z \in P_Z$ |
| $FullyConnected(1024) \Rightarrow BN(\cdot) \Rightarrow ReLU(\cdot)$ |
| $FullyConnected(128 \times \frac{h}{4} \times \frac{h}{4}) \Rightarrow BN(\cdot) \Rightarrow ReLU(\cdot)$ |
| $DeConv(64) \Rightarrow BN(\cdot) \Rightarrow ReLU(\cdot)$ |
| $DeConv(ch)$ |
| $Sigmoid(\cdot)$ |

(a)

|  |
|---|
| **Input:** $x_{U/P} \in X_{U/P}$ |
| $Conv(64) \Rightarrow SN(\cdot) \Rightarrow LeakyReLU(\cdot)$ |
| $Conv(128) \Rightarrow SN(\cdot) \Rightarrow LeakyReLU(\cdot)$ |
| $FullyConnected(1024) \Rightarrow SN(\cdot) \Rightarrow LeakyReLU(\cdot)$ |
| $Dropout(\cdot)$ |
| $FullyConnected(1)$ |
| $Sigmoid(\cdot)$ |

(b)

Figure 2: (a) is the generator network used for MNIST, Fashion-MNIST and CIFAR-10 datasets. $h = 28$ and $ch = 1$ for MNIST and Fashion-MNIST datasets, and $h = 32$ and $ch = 3$ for CIFAR-10 dataset. (b) is the classifier used for the discriminator and observer network for the three datasets. Size of $z$ is 100. Kernel size in all convolution layers is $4 \times 4$. We use transposed convolutional layers (DeConv) and Batch Normalization (BN) in the generator network. In classifier networks, we use convolutional layers with Spectral Normalization (SN).

same architecture as that used for the discriminator network of the GAN. We trained the second stage classifier for 1000 epochs, and picked the best performing model.

**Baselines**

We compare the performance of the $Observer$ network to *D-GAN*, which trains a binary classifier on the positive sample and previously generated negative samples, and ($TED^n$), which uses an iterative method between estimating $\alpha$ and training a binary classifier.

Because of the nature of the problem, as explained previously, the use of a validation set to choose the best learner is not feasible. We propose to use the Fréchet Inception Distance (FID) score (Heusel et al. 2017) as a tool to monitor training. Since the FID score accounts for both the quality and diversity of the generated images, it is a reasonable way to ensure that mode collapse does not happening during training. Monitoring the FID score of the generated images together with reinitialization, allows us to avoid mode collapse and overfitting of the $Observer$ network. As a result, the $Observer$ network tends to improve with training for more epochs. To evaluate the validity of this claim, we train all models in all of the experiments for 1000 epochs, and assess the average performance of the $Observer$ network on the test set using the last 50 and 100 epochs. We follow the same approach when training ($TED^n$). However, for *D-GAN*, the second step classifier can not avoid the overfitting problem, hence, we use a fully-labeled validation set to pick the best performing classifier to compare to. We train the $GAN$ model in *D-GAN* for 1000 epochs, and we train a classifier for 1000 epochs on the output of the $GAN$ after 100, 200, and 1000 epochs.

Table 2 shows the positive-versus-negative classification accuracy of each of the methods, applied on each of the datasets. The second step classifier of *D-GAN* fails for more complicated datasets since it learns only the difference between the real and fake samples and therefore predicts a positive label for all unseen realistic data. *D-GAN* appears to work well only for datasets that can easily be generated, which is not always the case in real-world PU datasets.

The *Observer-GAN* shows the best performance using the average accuracy of last 50 and 100 epochs. The performance of *Observer-GAN* improves from the last 100 to the last 50 in terms of average accuracy, which empirically supports our claim about continuous improvement of *Observer-GAN*. We can see in Figure 4 the test accuracy of the $Observer$ network and the FID score of the generated samples and the unlabeled and positive sets against the number of training epochs. We see that the reinitialization trick impacts the testing accuracy after each 100 epochs, yet the accuracy returns back to the highest after each drop.

Figure 5 shows sample output images from the generator of both *Observer-GAN* and *D-GAN*. It is evident that, although image quality is limited in both cases, *Observer-GAN* performs better in terms of quality of generated images, and enjoys more diversity in the output images in the later training epochs.

## Discussion

A human being can classify between cats and dogs with reasonable accuracy by looking only at a specific features (e.g. the tail). In general, neural networks are trained to classify between two classes of images in a similar way, i.e., given a labeled training set, the neural network is trained to highlight the important parts of the images concerning the classification task. Looking at Figure 5, it is easy to see that all the generated images from the generator of *Observer-GAN* contain a noticeable amount of noise, however, the $Observer$ network can still achieve good accuracy on the unseen test set. Because of the noisy output of any fixed generator network, a classifier trained with the positive dataset (cat images) and the output of a fixed generator network (fake dog images) as the input, learns the noise in the fake samples as it provides more evident differences between the two input classes. This holds for the output of any fixed generator in both *Observer-GAN* and *D-GAN*. However, in *Observer-GAN*, the learning is not done on a fixed generator network, which leads the $Observer$ network to only learn the features of the images that the generators are consistently generating (this can be seen in Figure 5 where the nose and eyes of each dog are persistent through all generator networks, which are unique features that distinguish dogs form cats), rather than the remaining, consistently-changing parts of the output images.

|                                  | D-GAN       | $TED^n$        |                 | Observer       |                |
|----------------------------------|-------------|----------------|-----------------|----------------|----------------|
|                                  | Early Stop  | 50             | 100             | 50             | 100            |
| **AFHQ (Cats vs. Dogs)**         | $\sim 50$   | $86.8 \pm 12$  | $89.9 \pm 14.9$ | $91 \pm 1.1$   | $90.1 \pm 3.2$ |
| **CIFAR (Animal vs. Not Animal)**| 82          | $88 \pm 2.5$   | $87.7 \pm 4.6$  | $89.6 \pm 0.7$ | $88.8 \pm 1.7$ |
| **MNIST (Even vs. Odd)**         | 98.3        | $97.7 \pm 0.4$ | $97.7 \pm 0.4$  | $98.3 \pm 0.2$ | $97.8 \pm 1.6$ |
| **Binarized Fashion MNIST**      | 89.6        | $88.5 \pm 0.9$ | $88.1 \pm 1$    | $92.6 \pm 0.3$ | $92 \pm 1$     |

Figure 3: (a) is the generator network used for AFHQ dataset. (b) is the classifier used for both the discriminator and observer network for AFHQ dataset. Size of $z$ is 100. Kernel size in all convolution layers is $5 \times 5$.

Table 2: Details of the experiments; Left-most column is the dataset, and upper-most row is the method used. We report the mean and standard deviation of the accuracy(%) of the last 50 and 100 epochs when using $TED^n$ or $Observer$ network, and the best performing model when using *D-GAN*.

Using the FID score to monitor training allows for identification of a mode collapse. In Figure 4, the FID scores appear to start increasing in the case of a mode collapse, likely because of the lack of diversity of the output of a collapsed GAN. We can see in the figure that this does not happen when training the *Observer-GAN*. While we expected to see the FID score between the generated samples and the unlabeled set to be lower in the later training stages than the FID score from the positive set, because the generator is trained to generate negative samples that are exclusively existent in the unlabeled set, we see that this is not evidently captured by the FID score. This might be caused by the fact that the FID score assesses the quality of the whole generated images, which have a considerable amount of noise. Therefore the difference between the generated samples and both the unlabeled and positive images is mostly represented by the noisy parts of the images. This also gives a better intuition about why a second stage classifier cannot be trained on the output of the GAN. Looking at the FID score of the generated images, we see stability (or a small decrease), and this reflects stability (or a small increase) of the test accuracy of the $Observer$ network as training progresses.

While Table 2 shows that $TED^n$ has comparable performance to the $Observer$ network for all of the datasets, it is worth noting that in $TED^n$ training, all of the positive and unlabeled datasets are used, on the other hand, after training the $Observer$ network, the final classifier will have only seen the positive samples. This leaves substantial room for improvement by making use of the unlabeled dataset for fine tuning the $Observer$ network. The $Observer$ network can generate pseudo-labels for the unlabeled data samples. The unlabeled data samples along with their corresponding pseudo-labels can then represent a new input dataset for learning methods that are robust to noisy labels (Song et al. 2022).

## Conclusion

In this paper, we present *Observer-GAN*, a method for learning from positive and unlabeled datasets. It employs a $GAN$ architecture with two discriminators. One of the discriminators (the $Observer$) contributes to generator learning by deviating its output from the positive class distribution, and it is also used as a final stage classifier between positive and negative testing samples. Because the $Observer$ network can give state-of-the-art classification accuracy while it does not use the unlabeled dataset as a direct input, a future direction to improve its performance is to apply a second-stage fine tuning method that is robust to noise and uses the unlabeled data samples and their corresponding pseudo labels(Cui et al. 2022) that can be generated by the $Observer$ network after training the *Observer-GAN*.

We have found that a GAN learns more slowly to generate realistic-looking images when trained with two discriminators. However, we demonstrate that the $Observer$ network can learn to distinguish between the positive and negative distributions even without having to generate realistic-looking images. This is because the loss function of the gen-erator network drives the generated images to contain features that can be used by the two discriminators. These features resemble the difference between positive and negative images, and are generated before the images look realistic to the human eye.

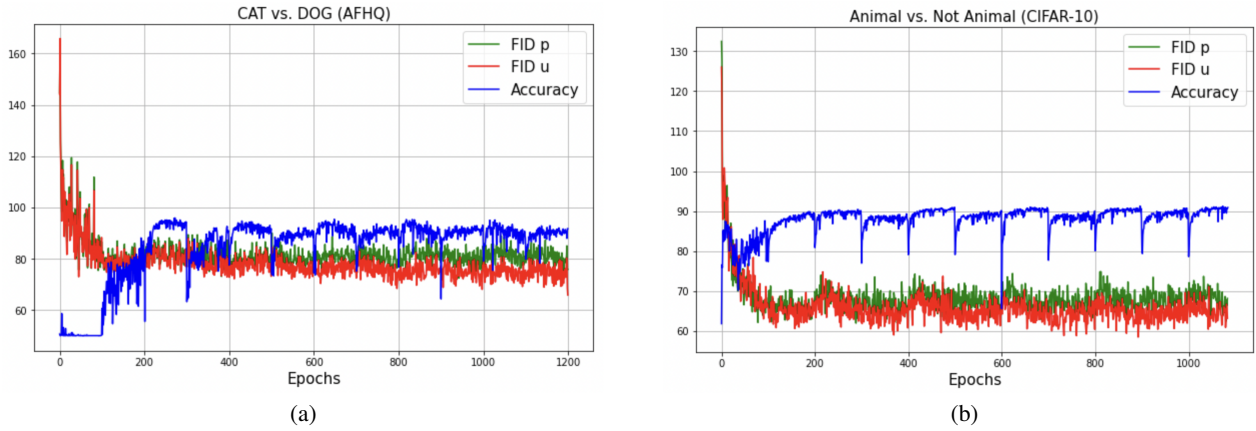

Figure 4: In blue, is the test accuracy curve of the *Observer* network as function of the training epochs. In red and green, is the FID score between the generated samples and the unlabeled data and positive data, respectively. (a) AFHQ, (b) CIFAR-10.

| Epoch | 200 | 400 | 600 | 800 | 1000 |
|---|---|---|---|---|---|
| Observer GAN | | | | | |
| Observer Accuracy | 81.1% | 86.3% | 90.7% | 91.5% | 91% |
| D-GAN | | | | | |
| Real Images | | | | | |

Figure 5: Example images generated by the generator of *Observer-GAN* and *D-GAN*. Each column contains a minibatch of 2 randomly generated images from each method after the number of epochs indicated in the first row. The third row indicates the testing accuracy of the observer network after the corresponding number of epochs. The bottom row contains 5 example images from the training set.

We also address the issue of overfitting that most existing PU learning methods suffer from because of the absence of a labeled validation set. We show that the *Observer* network does not tend to memorize positive samples, but rather learns meaningful differences between the positive and negative (generated) samples.